\documentclass[preprint,12pt]{elsarticle}

\usepackage{amssymb}
\usepackage[table,xcdraw]{xcolor}
\usepackage{makecell}

\usepackage{hyperref}
\hypersetup{hidelinks}

\journal{International Journal of Medical Informatics }

\begin{document}

\begin{frontmatter}

\title{Explainable Machine Learning for ICU Readmission Prediction}

\author[label1,label2,label3,label4]{Alex G. C. de Sá\corref{cor1}\fnref{cor2}}
\author[label7,label8]{Daniel Gould\fnref{cor2}}
\author[label5]{Anna Fedyukova\fnref{cor2}}

\author[label9,label10]{Mitchell Nicholas}
\author[label11]{Lucy Dockrell}
\author[label11,label11b]{Calvin Fletcher}

\author[label11,label12,label13]{David Pilcher}
\author[label5]{Daniel Capurro}
\author[label1,label2,label3,label4]{David B. Ascher\corref{cor1}}
\author[label11,label14,label15]{Khaled El-Khawas\corref{cor1}\fnref{cor2}}
\author[label1,label2,label3,label5]{Douglas E. V. Pires\corref{cor1}}

\affiliation[label1]{organization={School of Chemistry and Molecular Biosciences, The University of Queensland},
            city={Brisbane City},
            state={Queensland},
            country={Australia}}
\affiliation[label2]{organization={Systems and Computational Biology, Bio21 Institute, The University of Melbourne},
            city={Parkville},
            state={Victoria},
            country={Australia}}

\affiliation[label3]{organization={Computational Biology and Clinical Informatics, Baker Heart and Diabetes Institute},
            city={Melbourne},
            state={Victoria},
            country={Australia}}
            
\affiliation[label4]{organization={Baker Department of Cardiometabolic Health, The University of Melbourne},
            city={Parkville},
            state={Victoria},
            country={Australia}}

\affiliation[label7]{organization={Faculty of Medicine, Dentistry and Health Sciences, The University of Melbourne},
            city={Parkville},
            state={Victoria},
            country={Australia}}      

\affiliation[label8]{organization={The Department of Surgery, St. Vincent’s Hospital Melbourne},
            city={Fitzroy},
            state={Victoria},
            country={Australia}}              

\affiliation[label5]{organization={Centre for the Digital Transformation of Health, School of Computing and Information Systems, The University of Melbourne},
            city={Parkville},
            state={Victoria},
            country={Australia}}

\affiliation[label9]{organization={The Department of Anaesthesia and Perioperative Medicine, Monash Health},
            city={Clayton},
            state={Victoria},
            country={Australia}}  
            
\affiliation[label10]{organization={The Intensive Care Unit at Peninsula Private Hospital, Ramsay Healthcare},
            city={Langwarrin},
            state={Victoria},
            country={Australia}} 

\affiliation[label11]{organization={The Department of Intensive Care Medicine, The Alfred Hospital},
            city={ Prahran},
            state={Victoria},
            country={Australia}}    

\affiliation[label11b]{organization={The Department of Anaesthesiology and Perioperative Medicine, The Alfred Hospital},
            city={Melbourne},
            state={Victoria},
            country={Australia}}             

\affiliation[label12]{organization={The Australian and New Zealand Intensive Care Society Centre for Outcome and Resource Evaluation},
            city={Camberwell},
            state={Victoria},
            country={Australia}}       

\affiliation[label13]{organization={The Australian and New Zealand Intensive Care Research Centre, School of Public Health and Preventive Medicine, Monash University},
            city={Melbourne},
            state={Victoria},
            country={Australia}}  

\affiliation[label14]{organization={The Department  of  Intensive  Care,  Ballarat  Base  Hospital},
            city={Ballarat},
            state={Victoria},
            country={Australia}} 

\affiliation[label15]{organization={The Department of Intensive Care, Austin Hospital},
            city={Heidelberg},
            state={Victoria},
            country={Australia}}       


\cortext[cor1]{To whom correspondence should be addressed to D.E.V.P. Telephone: +61 383 448 185; Email: douglas.pires@unimelb.edu.au. Correspondence may also be addressed to A.G.C.S. at alex.desa@baker.edu.au, D.B.A. at d.ascher@uq.edu.au or K.E-K. at k.elkhawas@alfred.org.au.}
\fntext[cor2]{These authors contributed equally to this work.}

\begin{abstract}

\paragraph{Background}
The intensive care unit (ICU) comprises a complex hospital environment, where decisions made by clinicians have a high level of risk for the patients’ lives. Uncertain, competing and unplanned aspects within the ICU environment increase the difficulty in uniformly implementing the care pathway. Readmission contributes to this pathway's difficulty, resulting in high mortality rates and resource utilisation.

\paragraph{Objectives}
 Several works have tried to predict readmission through patients’ medical information. Although they have some level of success while predicting readmission, those works do not properly assess, characterise and understand readmission prediction. This work proposes a standardised and explainable machine learning pipeline to model, predict and explain patient ICU readmission.
 
 \paragraph{Method}
 This work focuses on using machine learning to model readmission using a multicentric eICU database while validating it on both MIMIC-IV monocentric and eICU multicentric settings. SHAP (Shapley Additive Explanations), calibration and likelihood ratio analysis are employed to comprehend the behaviour of the predictive model.

\paragraph{Results}
Our machine learning pipeline achieved predictive performance in the area of the receiver operating characteristic curve (AUC) up to 0.7 with a Random Forest classification model, yielding an overall consistency and generalisation on validation sets. After an interpretation analysis of our model through likelihood ratios and calibration, we observed that the proposed methodology could generate predictive models with proper and translatable diagnosis capabilities. In addition, from explanations provided by the constructed model, we could also derive a set of insightful conclusions, primarily on variables related to vital signs and blood tests (e.g., albumin, blood urea nitrogen and haemoglobin levels), demographics (e.g., age, and admission height and weight), and ICU-associated variables (e.g., unit type). 

\paragraph{Conclusions}
The model's predictions, behaviour and explainability method provided insights that yielded valuable information, which clinicians might (be able to) use to make decisions while discharging ICU patients.

\end{abstract}



\end{frontmatter}

\section{Introduction}

Intensive care is a service for patients with likely recoverable and treatable conditions, providing them with more precise observation and invasive treatments \cite{Smith1999}. Appropriate patient management within an intensive care unit~(ICU) is crucial. In a given ICU, the likelihood of long-term length of stay~(LOS), organ failures, and mortality tends to increase if adequate management is not taken into consideration \cite{Moitra2016, Sakr2012, Ay2020, Zhang2024}. Clinicians who are caring for these patients also have multiple competing activities to consider~\cite{Sujan2015, Turnbull2018}, including the potential deterioration of the patient’s condition after discharge, emergency admissions, elective admissions for high-risk surgery, staffing and resources, and evidence for appropriate allocation of resources. These aspects highlight the challenges across an ICU setting.

Readmission is one of the factors that extends the challenges within an ICU setting. It occurs when patients are admitted again to the ICU in a short timeframe (between 48 hours and 30 days), resulting in high mortality rates, increased LOS and, consequently, high resource utilisation \cite{Zhang2024, Mcneill2020, Alban2006, Kramer2013}. In summary, when readmission occurs, it disrupts the care pathway for the patients and poses additional challenges to the clinical team and hospital caring for them.

Reducing ICU readmission rates might not only improve the care pathway, leading to better patient outcomes, but also affect the hospital’s bottom line \cite{James2013,Khera2018}. Predicting patients at high risk of readmission would not only allow early intervention to reduce the risk but also reduce mortality rates, reduce resource utilisation (based on LOS) and, potentially, hospital costs.

The emergence of abundant hospital electronic health record~(EHR) data \cite{Yadav2018,Kim2019,Shickel2017} allowed the use of machine learning (ML) models targeting ICU readmission \cite{Badawi2012,Rojas2018,Mohanty2022,Thoral2021,Mcwilliams2019,Hammer2021,Lin2019,Xue2019,Barbieri2020, Hegselmann2022,Ashfaq2019,Lv2023}. These models for general readmission prediction have been summarised in Table S1. Apart from their undeniable importance, these models lack proper interpretation, generalisation and validation in different ICU settings -- e.g., monocentric (one hospital centre) \emph{versus} multicentric (multiple ICU hospital centres).

Our main contention is that predicting ICU readmission in a well-generali-sable way with hospital-based datasets is still a challenge to overcome, even after taking into account these univariate predictive models and best practice ML techniques \cite{Lin2019, Hosein2013, Curto2016}. The main reason for this difficulty is that medical data is usually noisy, uncertain, and characterised by a high degree of missingness, as it relies on different human inputs. Further exacerbating this complexity is data imbalance, with only a small proportion of patients being readmitted. ICU patients who happen to be readmitted are also very heterogeneous \cite{Hosein2013}, meaning that finding novel patterns across an ICU readmission data cohort tends to be challenging. As a result, an appropriate translation from ML models to the clinical environment is still limited. To improve the clinical translation, databases (e.g., MIMIC-IV and eICU) should be homogenised and have their noise and missingness diminished.

Accordingly, this work proposes a standardised methodological pipeline for a comprehensive assessment of ICU readmission modelling, prediction and understanding~\cite{Pollard2018a, Pollard2018b} using machine learning models \cite{Mitchell1997,Raschka2022} and statistical analyses. The produced predictive ICU readmission model -- which was trained and internally validated in a dataset (i.e., eICU) containing several hospitals (multicentric) and externally validated in a monocentric dataset (i.e., MIMIC IV)~\cite{Pollard2018a,Pollard2018b,Johnson2021} -- shows clear signs of predictive generalisation. It is the first time in the literature that such an approach is employed to confirm the performance of an ICU readmission model. To the best of our knowledge, no other previous method has been validated in opposing ICU settings (i.e., monocentric and multicentric).
\section{Material and Methods}
\label{methods}

\begin{figure}
    \centering
    \includegraphics[scale=0.295]{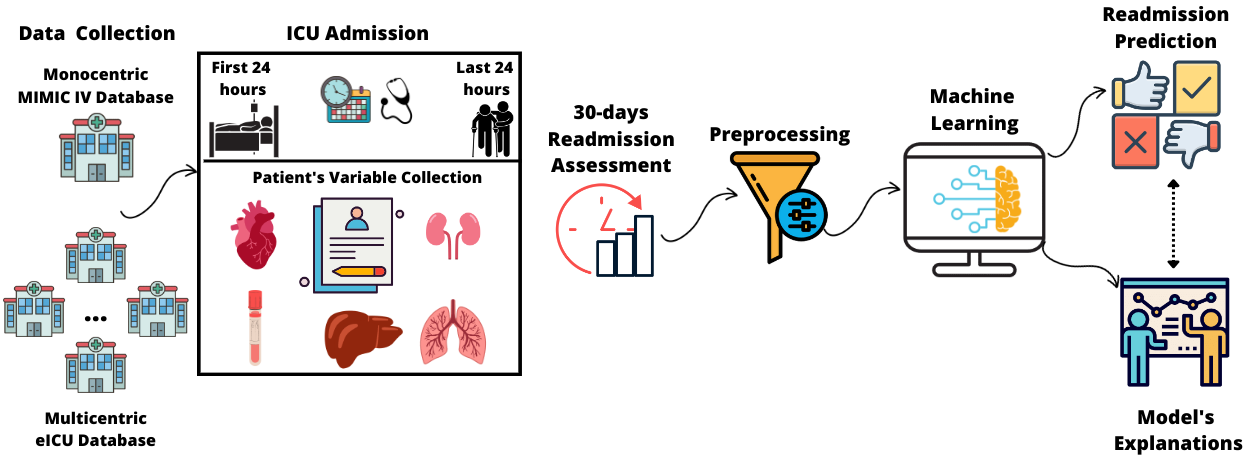}
    \caption{\textbf{The methodological workflow followed by this work.} Data comes from monocentric and multicentric databases (i.e., MIMIC IV and eICU, respectively), where the characterisation of ICU patients, hospitals, and ICU(s) takes place. Preprocessing filters out, standardises and imputes data for the development of a 30-day readmission machine learning model,  which is built and validated on eICU and tested on MIMIC-IV. This model is interpreted and also used to drive explanations from variables and predictions, potentially assisting and guiding clinicians while  treating and discharging new patients from the ICU.}
    \label{fig_1}
\end{figure}

Figure \ref{fig_1} presents the followed methodological workflow. The first step is to collect data from two databases: MIMIC IV (Medical Information Mart for Intensive Care, Version IV)~\cite{Johnson2021, Goldberger2000} and eICU (eICU Collaborative Research Database)~\cite{Pollard2018a,Pollard2018b,Goldberger2000}. From these databases, variables describing the patient's information are created considering their admission timeframe (see Figure S1). These variables include patient demographics, pre-admission information, vital signs and blood test results, and also variables outlining the hospitals and ICUs. These variables are fully described in Tables S2 and S3.

Data preprocessing considers several steps, e.g., imputation, imbalance learning, and feature standardisation. Exclusion criteria are also taken into consideration to avoid abnormal patient information in our dataset (see Figure S2). Machine learning algorithms take these preprocessed features as input to build a readmission model based on a timeframe of 30 days. Hyperparameter optimisation also occurs (Table S4), guaranteeing the selection of the most appropriate predictive model for the data at hand. The final employed machine learning algorithms use multicentric data from eICU to build and internally validate the predictive model for readmission while externally testing and validating the final model on monocentric data from MIMIC-IV. The readmission model predicts whether a given patient will be readmitted or not and supports explanations for its predictions, driving insights to clinicians while discharging a given patient in the ICU.

The Supplementary Material (more specifically, Supplementary Methods) provides complete details about data sources (eICU, MIMIC IV, and readmission outcome), predictor variables, data preprocessing, machine learning modelling and validation, and the model's explanation, diagnosis expression, and calibration.

\section{Results}

\subsection{eICU Readmission Group Characteristics, Outcomes and Comparison to MIMIC IV}

Table 1 provides the analyses contrasting readmitted and non-readmitted population groups. Out of 149,009 ICU admissions in the eICU database, readmission happens in 6,021 (4.04\%) patients within 30 days of their first ICU admission. We have observed that the readmitted group includes older, sicker and more male patients. These results are corroborated by a statistical test with a significance level lower than 1\%. The readmitted group also had a high level of gastrointestinal, respiratory and sepsis admission diagnoses when compared to the non-readmitted group (with a p-value $<$  1\%).

The combined medical-surgical ICU and cardiothoracic surgical ICU types reached the lowest proportion of readmissions, with statistical significance. The readmitted group received more vasopressors and renal replacement therapies when contrasted with the non-readmitted group. The readmitted group commonly had a longer ICU stay and, consequently, a longer hospital length of stay (p-value $<$  1\%). Finally, they were more than twice as likely to die in a hospital (16.0\% versus 7.1\% with a p-value $<$  1\%), with survivors needing more rehabilitation, home nursing or skilled nursing facilities (p-value $<$  1\%).

The Medical-Surgical Intensive Care Unit (MSICU) had the highest readmission rate at 49.1\%, followed by the Medical Intensive Care Unit (MICU) with 10.9\% and the Critical Care Unit-Cardio-Thoracic Intensive Care Unit (CCU-CTICU) with 10.1\%. However, the absolute difference between readmission and non-readmission rates in most ICU types does not exceed 2.7\%, except for MSICU, with a 6.6\% difference between the two groups. The lowest readmission rate has been found in the Cardio-Surgical Intensive Care Unit (CSICU).

\begin{table}[]
\centering
\caption{Characteristics of discharged readmitted and non-readmitted patients in eICU.}
\tiny
\begin{tabular}{lllll}
\cline{1-4}
\multicolumn{1}{|l|}{\cellcolor[HTML]{FFFFFF}\textbf{Characteristic}}                                                                 & \multicolumn{1}{l|}{\cellcolor[HTML]{FFFFFF}\textbf{\begin{tabular}[c]{@{}l@{}}Readmitted\\ (value with \% OR \\ mean/median)\end{tabular}}} & \multicolumn{1}{l|}{\cellcolor[HTML]{FFFFFF}\textbf{\begin{tabular}[c]{@{}l@{}}Non-Readmitted\\ (value with \% OR \\ mean/median)\end{tabular}}} & \multicolumn{1}{l|}{\cellcolor[HTML]{FFFFFF}\textbf{p-value}}   &  \\ \cline{1-4}
\multicolumn{1}{|l|}{\cellcolor[HTML]{FFFFFF}\textbf{Readmission Numbers}}                                                            & \multicolumn{1}{l|}{\cellcolor[HTML]{FFFFFF}6,021}                                                                                        & \multicolumn{1}{l|}{\cellcolor[HTML]{FFFFFF}142,988}                                                                                          & \multicolumn{1}{c|}{\cellcolor[HTML]{FFFFFF}-}                  &  \\ \cline{1-4}

\multicolumn{4}{l}{}                                                                                                                                                                                                                                                                                                                   &  \\ \cline{1-4}
\multicolumn{4}{|l|}{\cellcolor[HTML]{FFFFFF}\textbf{Gender:}}                                                                                                                                                                                                                                                                                                                                                                                                                                      &  \\ \cline{1-4}
\multicolumn{1}{|l|}{\cellcolor[HTML]{FFFFFF}Male}                                                                                    & \multicolumn{1}{l|}{\cellcolor[HTML]{FFFFFF}3,417 (56.8\%)}                                                                               & \multicolumn{1}{l|}{\cellcolor[HTML]{FFFFFF}77,144 (54.0\%)}                                                                                  & \multicolumn{1}{l|}{\cellcolor[HTML]{FFFFFF}2.198e-05*}         &  \\ \cline{1-4}
\multicolumn{1}{|l|}{\cellcolor[HTML]{FFFFFF}Female}                                                                                  & \multicolumn{1}{l|}{\cellcolor[HTML]{FFFFFF}2,602 (43.2\%)}                                                                               & \multicolumn{1}{l|}{\cellcolor[HTML]{FFFFFF}65,771 (46.0\%)}                                                                                  & \multicolumn{1}{l|}{\cellcolor[HTML]{FFFFFF}2.198e-05*}         &  \\ \cline{1-4}
\multicolumn{4}{l}{}                                                                                                                                                                                                                                                                                                                                                                                                                                                                                &  \\ \cline{1-4}
\multicolumn{1}{|l|}{\cellcolor[HTML]{FFFFFF}\textbf{Age}}                                                                            & \multicolumn{1}{l|}{\cellcolor[HTML]{FFFFFF}65.48 / 67.00}                                                                                & \multicolumn{1}{l|}{\cellcolor[HTML]{FFFFFF}62.70 / 65.00}                                                                                    & \multicolumn{1}{l|}{\cellcolor[HTML]{FFFFFF}\textless 2.2e-16*} &  \\ \cline{1-4}
\multicolumn{1}{|l|}{\cellcolor[HTML]{FFFFFF}\textbf{Body Mass Index (BMI)}}                                                          & \multicolumn{1}{l|}{\cellcolor[HTML]{FFFFFF}28.69 / 27.30}                                                                                & \multicolumn{1}{l|}{\cellcolor[HTML]{FFFFFF}29.05 / 27.51}                                                                                    & \multicolumn{1}{l|}{\cellcolor[HTML]{FFFFFF}\textless 0.001*}   &  \\ \cline{1-4}
\multicolumn{1}{|l|}{\cellcolor[HTML]{FFFFFF}\textbf{APACHE IV}}                                                                      & \multicolumn{1}{l|}{\cellcolor[HTML]{FFFFFF}59.11 / 56.00}                                                                                & \multicolumn{1}{l|}{\cellcolor[HTML]{FFFFFF}53.86 / 50.00}                                                                                    & \multicolumn{1}{l|}{\cellcolor[HTML]{FFFFFF}\textless 2.2e-16*} &  \\ \cline{1-4}
\multicolumn{4}{l}{}                                                                                                                                                                                                                                                                                                                                                                                                                                                                                &  \\ \cline{1-4}
\multicolumn{4}{|l|}{\cellcolor[HTML]{FFFFFF}\textbf{APACHE Diagnosis:}}                                                                                                                                                                                                                                                                                                                                                                                                                            &  \\ \cline{1-4}
\multicolumn{1}{|l|}{\cellcolor[HTML]{FFFFFF}Cardiovascular}                                                                          & \multicolumn{1}{l|}{\cellcolor[HTML]{FFFFFF}1,789 (29.7\%)}                                                                               & \multicolumn{1}{l|}{\cellcolor[HTML]{FFFFFF}45,423 (31.8\%)}                                                                                  & \multicolumn{1}{l|}{\cellcolor[HTML]{FFFFFF}\textless 0.001*}   &  \\ \cline{1-4}
\multicolumn{1}{|l|}{\cellcolor[HTML]{FFFFFF}Neurological}                                                                            & \multicolumn{1}{l|}{\cellcolor[HTML]{FFFFFF}881 (14.6\%)}                                                                                 & \multicolumn{1}{l|}{\cellcolor[HTML]{FFFFFF}27,068 (18.9\%)}                                                                                  & \multicolumn{1}{l|}{\cellcolor[HTML]{FFFFFF}\textless 2.2e-16*} &  \\ \cline{1-4}
\multicolumn{1}{|l|}{\cellcolor[HTML]{FFFFFF}Gastrointestinal}                                                                        & \multicolumn{1}{l|}{\cellcolor[HTML]{FFFFFF}798 (13.3\%)}                                                                                 & \multicolumn{1}{l|}{\cellcolor[HTML]{FFFFFF}13,755 (9.6\%)}                                                                                   & \multicolumn{1}{l|}{\cellcolor[HTML]{FFFFFF}\textless 2.2e-16*} &  \\ \cline{1-4}
\multicolumn{1}{|l|}{\cellcolor[HTML]{FFFFFF}Trauma}                                                                                  & \multicolumn{1}{l|}{\cellcolor[HTML]{FFFFFF}230 (3.8\%)}                                                                                  & \multicolumn{1}{l|}{\cellcolor[HTML]{FFFFFF}6,470 (4.5\%)}                                                                                    & \multicolumn{1}{l|}{\cellcolor[HTML]{FFFFFF}0.011}              &  \\ \cline{1-4}
\multicolumn{1}{|l|}{\cellcolor[HTML]{FFFFFF}Respiratory}                                                                             & \multicolumn{1}{l|}{\cellcolor[HTML]{FFFFFF}1,045 (17.4\%)}                                                                               & \multicolumn{1}{l|}{\cellcolor[HTML]{FFFFFF}22,158 (15.5\%)}                                                                                  & \multicolumn{1}{l|}{\cellcolor[HTML]{FFFFFF}\textless 0.001*}   &  \\ \cline{1-4}
\multicolumn{1}{|l|}{\cellcolor[HTML]{FFFFFF}Sepsis}                                                                                  & \multicolumn{1}{l|}{\cellcolor[HTML]{FFFFFF}972 (16.1\%)}                                                                                 & \multicolumn{1}{l|}{\cellcolor[HTML]{FFFFFF}17,404 (12.2\%)}                                                                                  & \multicolumn{1}{l|}{\cellcolor[HTML]{FFFFFF}\textless 2.2e-16*} &  \\ \cline{1-4}
\multicolumn{1}{|l|}{\cellcolor[HTML]{FFFFFF}Other}                                                                                   & \multicolumn{1}{l|}{\cellcolor[HTML]{FFFFFF}412 (6.8\%)}                                                                                  & \multicolumn{1}{l|}{\cellcolor[HTML]{FFFFFF}13,810 (9.7\%)}                                                                                   & \multicolumn{1}{l|}{\cellcolor[HTML]{FFFFFF}3.8e-13*}           &  \\ \cline{1-4}
\multicolumn{4}{l}{}                                                                                                                                                                                                                                                                                                                                                                                                                                                                                &  \\ \cline{1-4}
\multicolumn{4}{|l|}{\cellcolor[HTML]{FFFFFF}\textbf{Unit Type:}}                                                                                                                                                                                                                                                                                                                                                                                                                                   &  \\ \cline{1-4}
\multicolumn{1}{|l|}{\cellcolor[HTML]{FFFFFF}\makecell[l]{Critical Care Unit-Cardio-Thoracic Intensive \\ Care Unit (CCU-CTICU)}}                      & \multicolumn{1}{l|}{\cellcolor[HTML]{FFFFFF}610 (10.1\%)}                                                                                 & \multicolumn{1}{l|}{\cellcolor[HTML]{FFFFFF}12,322 (8.6\%)}                                                                                   & \multicolumn{1}{l|}{\cellcolor[HTML]{FFFFFF}4.8e-05*}           &  \\ \cline{1-4}
\multicolumn{1}{|l|}{\cellcolor[HTML]{FFFFFF}Cardio-Surgical Intensive Care Unit (CSICU)}                                             & \multicolumn{1}{l|}{\cellcolor[HTML]{FFFFFF}90 (1.5\%)}                                                                                   & \multicolumn{1}{l|}{\cellcolor[HTML]{FFFFFF}5,335 (3.7\%)}                                                                                    & \multicolumn{1}{l|}{\cellcolor[HTML]{FFFFFF}\textless 2.2e-16*} &  \\ \cline{1-4}
\multicolumn{1}{|l|}{\cellcolor[HTML]{FFFFFF}Cardio-Thoracic Intensive Care Unit (CTICU)}                                             & \multicolumn{1}{l|}{\cellcolor[HTML]{FFFFFF}291 (4.8\%)}                                                                                  & \multicolumn{1}{l|}{\cellcolor[HTML]{FFFFFF}4,750 (3.3\%)}                                                                                    & \multicolumn{1}{l|}{\cellcolor[HTML]{FFFFFF}2.7e-10*}           &  \\ \cline{1-4}
\multicolumn{1}{|l|}{\cellcolor[HTML]{FFFFFF}Cardiac Intensive Care Unit (CICU)}                                                      & \multicolumn{1}{l|}{\cellcolor[HTML]{FFFFFF}440 (7.3\%)}                                                                                  & \multicolumn{1}{l|}{\cellcolor[HTML]{FFFFFF}9,392 (6.6\%)}                                                                                    & \multicolumn{1}{l|}{\cellcolor[HTML]{FFFFFF}0.025}              &  \\ \cline{1-4}
\multicolumn{1}{|l|}{\cellcolor[HTML]{FFFFFF}Medical Intensive Care Unit (MICU)}                                                      & \multicolumn{1}{l|}{\cellcolor[HTML]{FFFFFF}657 (10.9\%)}                                                                                 & \multicolumn{1}{l|}{\cellcolor[HTML]{FFFFFF}11,710 (8.2\%)}                                                                                   & \multicolumn{1}{l|}{\cellcolor[HTML]{FFFFFF}7.6e-14*}           &  \\ \cline{1-4}
\multicolumn{1}{|l|}{\cellcolor[HTML]{FFFFFF}Surgical Intensive Care Unit (SICU)}                                                     & \multicolumn{1}{l|}{\cellcolor[HTML]{FFFFFF}503 (8.4\%)}                                                                                  & \multicolumn{1}{l|}{\cellcolor[HTML]{FFFFFF}8,854 (6.2\%)}                                                                                    & \multicolumn{1}{l|}{\cellcolor[HTML]{FFFFFF}1.5e-11*}           &  \\ \cline{1-4}
\multicolumn{1}{|l|}{\cellcolor[HTML]{FFFFFF}Medical-Surgical Intensive Care Unit (MSICU)}                                            & \multicolumn{1}{l|}{\cellcolor[HTML]{FFFFFF}2,955 (49.1\%)}                                                                               & \multicolumn{1}{l|}{\cellcolor[HTML]{FFFFFF}79,613 (55.7\%)}                                                                                  & \multicolumn{1}{l|}{\cellcolor[HTML]{FFFFFF}\textless 2.2e-16*} &  \\ \cline{1-4}
\multicolumn{1}{|l|}{\cellcolor[HTML]{FFFFFF}Neurological  Intensive Care Unit (NICU)}                                                & \multicolumn{1}{l|}{\cellcolor[HTML]{FFFFFF}475 (7.89\%)}                                                                                 & \multicolumn{1}{l|}{\cellcolor[HTML]{FFFFFF}11,012 (7.7\%)}                                                                                   & \multicolumn{1}{l|}{\cellcolor[HTML]{FFFFFF}0.610}              &  \\ \cline{1-4}
\multicolumn{4}{l}{}                                                                                                                                                                                                                                                                                                                                                                                                                                                                                &  \\ \cline{1-4}
\multicolumn{4}{|l|}{\cellcolor[HTML]{FFFFFF}\textbf{Hospital Capacity:}}                                                                                                                                                                                                                                                                                                                                                                                                                           &  \\ \cline{1-4}
\multicolumn{1}{|l|}{\cellcolor[HTML]{FFFFFF}\begin{tabular}[c]{@{}l@{}}Extra-Large Hospital \\ (\textgreater 500 beds)\end{tabular}} & \multicolumn{1}{l|}{\cellcolor[HTML]{FFFFFF}2,427 (40.3\%)}                                                                               & \multicolumn{1}{l|}{\cellcolor[HTML]{FFFFFF}51,538 (36.0\%)}                                                                                  & \multicolumn{1}{l|}{\cellcolor[HTML]{FFFFFF}1.7e-11*}           &  \\ \cline{1-4}
\multicolumn{1}{|l|}{\cellcolor[HTML]{FFFFFF}\begin{tabular}[c]{@{}l@{}}Large Hospital \\ (250 – 499 beds)\end{tabular}}              & \multicolumn{1}{l|}{\cellcolor[HTML]{FFFFFF}1,431 (23.8\%)}                                                                               & \multicolumn{1}{l|}{\cellcolor[HTML]{FFFFFF}33,108 (23.2\%)}                                                                                  & \multicolumn{1}{l|}{\cellcolor[HTML]{FFFFFF}0.277}              &  \\ \cline{1-4}
\multicolumn{1}{|l|}{\cellcolor[HTML]{FFFFFF}\begin{tabular}[c]{@{}l@{}}Medium-sized Hospital \\ (100 – 249 beds)\end{tabular}}       & \multicolumn{1}{l|}{\cellcolor[HTML]{FFFFFF}1,052 (17.5\%)}                                                                               & \multicolumn{1}{l|}{\cellcolor[HTML]{FFFFFF}31,539 (22.1\%)}                                                                                  & \multicolumn{1}{l|}{\cellcolor[HTML]{FFFFFF}\textless 2.2e-16*} &  \\ \cline{1-4}
\multicolumn{1}{|l|}{\cellcolor[HTML]{FFFFFF}\begin{tabular}[c]{@{}l@{}}Small hospital \\ (\textless 100 beds)\end{tabular}}          & \multicolumn{1}{l|}{\cellcolor[HTML]{FFFFFF}231 (3.8\%)}                                                                                  & \multicolumn{1}{l|}{\cellcolor[HTML]{FFFFFF}9,438 (6.6\%)}                                                                                    & \multicolumn{1}{l|}{\cellcolor[HTML]{FFFFFF}\textless 2.2e-16*} &  \\ \cline{1-4}
\multicolumn{1}{|l|}{\cellcolor[HTML]{FFFFFF}Unknown number of beds}                                                                  & \multicolumn{1}{l|}{\cellcolor[HTML]{FFFFFF}880 (14.6\%)}                                                                                 & \multicolumn{1}{l|}{\cellcolor[HTML]{FFFFFF}17,365 (12.1\%)}                                                                                  & \multicolumn{1}{l|}{\cellcolor[HTML]{FFFFFF}1.1e-08*}           &  \\ \cline{1-4}
\multicolumn{4}{l}{}                                                                                                                                                                                                                                                                                                                                                                                                                                                                                &  \\ \cline{1-4}
\multicolumn{4}{|l|}{\cellcolor[HTML]{FFFFFF}\textbf{Hospital Type:}}                                                                                                                                                                                                                                                                                                                                                                                                                               &  \\ \cline{1-4}
\multicolumn{1}{|l|}{\cellcolor[HTML]{FFFFFF}Teaching}                                                                                & \multicolumn{1}{l|}{\cellcolor[HTML]{FFFFFF}1,687 (28.0\%)}                                                                               & \multicolumn{1}{l|}{\cellcolor[HTML]{FFFFFF}35,715 (25.0\%)}                                                                                  & \multicolumn{1}{l|}{\cellcolor[HTML]{FFFFFF}1.1e-07*}           &  \\ \cline{1-4}
\multicolumn{1}{|l|}{\cellcolor[HTML]{FFFFFF}Non-teaching}                                                                            & \multicolumn{1}{l|}{\cellcolor[HTML]{FFFFFF}4,334 (72.0\%)}                                                                               & \multicolumn{1}{l|}{\cellcolor[HTML]{FFFFFF}107,273 (75.0\%)}                                                                                 & \multicolumn{1}{l|}{\cellcolor[HTML]{FFFFFF}1.1e-07*}           &  \\ \cline{1-4}
\multicolumn{4}{l}{}                                                                                                                                                                                                                                                                                                                                                                                                                                                                                &  \\ \cline{1-4}
\multicolumn{4}{|l|}{\cellcolor[HTML]{FFFFFF}\textbf{Patient Origin:}}                                                                                                                                                                                                                                                                                                                                                                                                                              &  \\ \cline{1-4}
\multicolumn{1}{|l|}{\cellcolor[HTML]{FFFFFF}ICU}                                                                                     & \multicolumn{1}{l|}{\cellcolor[HTML]{FFFFFF}152 (2.5\%)}                                                                                  & \multicolumn{1}{l|}{\cellcolor[HTML]{FFFFFF}3,261 (2.3\%)}                                                                                    & \multicolumn{1}{l|}{\cellcolor[HTML]{FFFFFF}0.232}              &  \\ \cline{1-4}
\multicolumn{1}{|l|}{\cellcolor[HTML]{FFFFFF}Operating Room}                                                                          & \multicolumn{1}{l|}{\cellcolor[HTML]{FFFFFF}564 (9.4\%)}                                                                                  & \multicolumn{1}{l|}{\cellcolor[HTML]{FFFFFF}14,675 (10.3\%)}                                                                                  & \multicolumn{1}{l|}{\cellcolor[HTML]{FFFFFF}0.026}              &  \\ \cline{1-4}
\multicolumn{1}{|l|}{\cellcolor[HTML]{FFFFFF}Emergency Department (ED)}                                                               & \multicolumn{1}{l|}{\cellcolor[HTML]{FFFFFF}2,089 (34.7\%)}                                                                               & \multicolumn{1}{l|}{\cellcolor[HTML]{FFFFFF}59,728 (41.8\%)}                                                                                  & \multicolumn{1}{l|}{\cellcolor[HTML]{FFFFFF}\textless 2.2e-16*} &  \\ \cline{1-4}
\multicolumn{1}{|l|}{\cellcolor[HTML]{FFFFFF}Recovery Room}                                                                           & \multicolumn{1}{l|}{\cellcolor[HTML]{FFFFFF}222 (3.7\%)}                                                                                  & \multicolumn{1}{l|}{\cellcolor[HTML]{FFFFFF}4,929 (3.4\%)}                                                                                    & \multicolumn{1}{l|}{\cellcolor[HTML]{FFFFFF}0.3359}             &  \\ \cline{1-4}
\multicolumn{1}{|l|}{\cellcolor[HTML]{FFFFFF}Unknown}                                                                                 & \multicolumn{1}{l|}{\cellcolor[HTML]{FFFFFF}1,545 (25.7\%)}                                                                               & \multicolumn{1}{l|}{\cellcolor[HTML]{FFFFFF}33,334 (23.3\%)}                                                                                  & \multicolumn{1}{l|}{\cellcolor[HTML]{FFFFFF}2.7e-05*}           &  \\ \cline{1-4}
\multicolumn{1}{|l|}{\cellcolor[HTML]{FFFFFF}Other}                                                                                   & \multicolumn{1}{l|}{\cellcolor[HTML]{FFFFFF}1,449 (24.1\%)}                                                                               & \multicolumn{1}{l|}{\cellcolor[HTML]{FFFFFF}27,061 (18.9\%)}                                                                                  & \multicolumn{1}{l|}{\cellcolor[HTML]{FFFFFF}\textless 2.2e-16*} &  \\ \cline{1-4}
\multicolumn{4}{l}{}                                                                                                                                                                                                                                                                                                                                                                                                                                                                                &  \\ \cline{1-4}
\multicolumn{4}{|l|}{\cellcolor[HTML]{FFFFFF}\textbf{Therapies in ICU:}}                                                                                                                                                                                                                                                                                                                                                                                                                            &  \\ \cline{1-4}
\multicolumn{1}{|l|}{\cellcolor[HTML]{FFFFFF}Mechanical Ventilation}                                                                  & \multicolumn{1}{l|}{\cellcolor[HTML]{FFFFFF}1,252 (20.8\%)}                                                                               & \multicolumn{1}{l|}{\cellcolor[HTML]{FFFFFF}27,651 (19.3\%)}                                                                                  & \multicolumn{1}{l|}{\cellcolor[HTML]{FFFFFF}0.005}              &  \\ \cline{1-4}
\multicolumn{1}{|l|}{\cellcolor[HTML]{FFFFFF}Vasopressors}                                                                            & \multicolumn{1}{l|}{\cellcolor[HTML]{FFFFFF}761 (12.6\%)}                                                                                 & \multicolumn{1}{l|}{\cellcolor[HTML]{FFFFFF}16,060 (11.2\%)}                                                                                  & \multicolumn{1}{l|}{\cellcolor[HTML]{FFFFFF}\textless 0.001*}   &  \\ \cline{1-4}
\multicolumn{1}{|l|}{\cellcolor[HTML]{FFFFFF}Renal Replacement Therapies (RRT)}                                                       & \multicolumn{1}{l|}{\cellcolor[HTML]{FFFFFF}232 (3.9\%)}                                                                                  & \multicolumn{1}{l|}{\cellcolor[HTML]{FFFFFF}4,406 (03.1\%)}                                                                                   & \multicolumn{1}{l|}{\cellcolor[HTML]{FFFFFF}\textless 0.001*}   &  \\ \cline{1-4}
\multicolumn{4}{l}{}                                                                                                                                                                                                                                                                                                                                                                                                                                                                                &  \\ \cline{1-4}
\multicolumn{4}{|l|}{\cellcolor[HTML]{FFFFFF}\textbf{Length of Stay (LOS):}}                                                                                                                                                                                                                                                                                                                                                                                                                        &  \\ \cline{1-4}
\multicolumn{1}{|l|}{\cellcolor[HTML]{FFFFFF}ICU LOS, days}                                                                           & \multicolumn{1}{l|}{\cellcolor[HTML]{FFFFFF}3.72 / 2.18}                                                                                  & \multicolumn{1}{l|}{\cellcolor[HTML]{FFFFFF}2.86 / 1.75}                                                                                      & \multicolumn{1}{l|}{\cellcolor[HTML]{FFFFFF}\textless 2.2e-16*} &  \\ \cline{1-4}
\multicolumn{1}{|l|}{\cellcolor[HTML]{FFFFFF}Hospital LOS, days}                                                                      & \multicolumn{1}{l|}{\cellcolor[HTML]{FFFFFF}17.86 / 14.64}                                                                                & \multicolumn{1}{l|}{\cellcolor[HTML]{FFFFFF}6.92 / 5.00}                                                                                      & \multicolumn{1}{l|}{\cellcolor[HTML]{FFFFFF}\textless 2.2e-16*} &  \\ \cline{1-4}
\multicolumn{4}{l}{}                                                                                                                                                                                                                                                                                                                                                                                                                                                                                &  \\ \cline{1-4}
\multicolumn{1}{|l|}{\cellcolor[HTML]{FFFFFF}\textbf{Hospital Mortality}}                                                             & \multicolumn{1}{l|}{\cellcolor[HTML]{FFFFFF}965 (16.0\%)}                                                                                 & \multicolumn{1}{l|}{\cellcolor[HTML]{FFFFFF}10,162 (7.1\%)}                                                                                   & \multicolumn{1}{l|}{\cellcolor[HTML]{FFFFFF}\textless 2.2e-16*} &  \\ \cline{1-4}
\multicolumn{4}{l}{}                                                                                                                                                                                                                                                                                                                                                                                                                                                                                &  \\ \cline{1-4}
\multicolumn{4}{|l|}{\cellcolor[HTML]{FFFFFF}\textbf{Discharge destination:}}                                                                                                                                                                                                                                                                                                                                                                                                                       &  \\ \cline{1-4}
\multicolumn{1}{|l|}{\cellcolor[HTML]{FFFFFF}Home}                                                                                    & \multicolumn{1}{l|}{\cellcolor[HTML]{FFFFFF}2,290 (38.0\%)}                                                                               & \multicolumn{1}{l|}{\cellcolor[HTML]{FFFFFF}87,841 (61.4\%)}                                                                                  & \multicolumn{1}{l|}{\cellcolor[HTML]{FFFFFF}\textless 2.2e-16*} &  \\ \cline{1-4}
\multicolumn{1}{|l|}{\cellcolor[HTML]{FFFFFF}Care Facility}                                                                           & \multicolumn{1}{l|}{\cellcolor[HTML]{FFFFFF}2,409 (40.0\%)}                                                                               & \multicolumn{1}{l|}{\cellcolor[HTML]{FFFFFF}38,220 (26.7\%)}                                                                                  & \multicolumn{1}{l|}{\cellcolor[HTML]{FFFFFF}\textless 2.2e-16*} &  \\ \cline{1-4}
\multicolumn{4}{l}{}                                                                                                                                                                                                                                                                                                                                                                                                                                                                                &  \\ \cline{1-4}
\multicolumn{4}{|l|}{\cellcolor[HTML]{FFFFFF}\textbf{Care Facility (Discharge destination):}}                                                                                                                                                                                                                                                                                                                                                                                                       &  \\ \cline{1-4}
\multicolumn{1}{|l|}{\cellcolor[HTML]{FFFFFF}Skilled Nursing Facility}                                                                & \multicolumn{1}{l|}{\cellcolor[HTML]{FFFFFF}1,150 (19.1\%)}                                                                               & \multicolumn{1}{l|}{\cellcolor[HTML]{FFFFFF}18,578 (13.0\%)}                                                                                  & \multicolumn{1}{l|}{\cellcolor[HTML]{FFFFFF}\textless 2.2e-16*} &  \\ \cline{1-4}
\multicolumn{1}{|l|}{\cellcolor[HTML]{FFFFFF}Other Hospital}                                                                          & \multicolumn{1}{l|}{\cellcolor[HTML]{FFFFFF}374 (6.2\%)}                                                                                  & \multicolumn{1}{l|}{\cellcolor[HTML]{FFFFFF}5,698 (4.0\%)}                                                                                    & \multicolumn{1}{l|}{\cellcolor[HTML]{FFFFFF}\textless 2.2e-16*} &  \\ \cline{1-4}
\multicolumn{1}{|l|}{\cellcolor[HTML]{FFFFFF}Rehabilitation}                                                                          & \multicolumn{1}{l|}{\cellcolor[HTML]{FFFFFF}428 (7.1\%)}                                                                                  & \multicolumn{1}{l|}{\cellcolor[HTML]{FFFFFF}6,213 (4.3\%)}                                                                                    & \multicolumn{1}{l|}{\cellcolor[HTML]{FFFFFF}\textless 2.2e-16*} &  \\ \cline{1-4}
\multicolumn{1}{|l|}{\cellcolor[HTML]{FFFFFF}Nursing Home}                                                                            & \multicolumn{1}{l|}{\cellcolor[HTML]{FFFFFF}128 (2.1\%)}                                                                                  & \multicolumn{1}{l|}{\cellcolor[HTML]{FFFFFF}1,604 (1.1\%)}                                                                                    & \multicolumn{1}{l|}{\cellcolor[HTML]{FFFFFF}1.7e-12*}           &  \\ \cline{1-4}
\multicolumn{1}{|l|}{\cellcolor[HTML]{FFFFFF}Other External}                                                                          & \multicolumn{1}{l|}{\cellcolor[HTML]{FFFFFF}329 (5.5\%)}                                                                                  & \multicolumn{1}{l|}{\cellcolor[HTML]{FFFFFF}6,127 (4.3\%)}                                                                                    & \multicolumn{1}{l|}{\cellcolor[HTML]{FFFFFF}1.2e-05*}           &  \\ \cline{1-4}
$\star$ Presence of statistical significance.
\end{tabular}
\end{table}

Figure S3 also highlights patients readmitted to another type of ICU where the patient was originally admitted.  Several ICUs in the eICU database have more than 30\% readmissions to other ICU types, such as the Surgical Intensive Care Unit (SICU), Medical Intensive Care Unit (MICU) and Cardiac Intensive Care Unit (CICU).

While aiming to compare MIMIC IV and eICU readmitted patients, we noticed several differences in the level of information these cohorts provide, including patients who do not have length of stay (LOS) and discharge location. As a result, MIMIC IV and eICU were reduced from 5,984 and 6,021 to 5,980 and 5,403, respectively. This step was performed only to compare both cohorts directly. Matching the characteristics of MIMIC IV and eICU is not straightforward overall. MIMIC IV does not include several characteristics contained in eICU, and vice-versa, meaning that the comparison across the readmission groups from eICU and MIMIC IV was restricted to only a few characteristics.

Apart from that, we could observe they have similar baseline characteristics (Table 2). However, MIMIC IV readmitted patients received more ventilation and vasopressors but fewer renal replacement therapies than the readmitted eICU patients. In addition, MIMIC IV readmitted patients stayed longer in the ICU and hospital stay. Mortality was also slightly higher in the MIMIC IV readmission patients, although no statistical difference has been found between the two databases. Among survivors, the readmission group in eICU was discharged more to home, while readmitted patients in MIMIC IV had a higher rate of further treatments in a rehabilitation care facility (p-value $<$ 1\%).

\begin{table}[]
\centering
\caption{Comparisons of characteristics between eICU and MIMIC IV readmission groups.}
\tiny
\begin{tabular}{lllll}
\cline{1-4}
\multicolumn{1}{|l|}{\textbf{Characteristic}}                  & \multicolumn{1}{l|}{\textbf{\begin{tabular}[c]{@{}l@{}}eICU\\ (value with \% OR \\ mean/median)\end{tabular}}} & \multicolumn{1}{l|}{\textbf{\begin{tabular}[c]{@{}l@{}}MIMIC \\ (value with \% OR \\  mean/median)\end{tabular}}} & \multicolumn{1}{l|}{\textbf{p-value}}   &  \\ \cline{1-4}
\multicolumn{1}{|l|}{\textbf{Readmission Numbers}}             & \multicolumn{1}{l|}{5,403}                                                                                  & \multicolumn{1}{l|}{5,980}                                                                                    & \multicolumn{1}{l|}{-}                  &  \\ \cline{1-4}
\multicolumn{4}{l}{}                                                                                                                                                                                                                                                                                                                   &  \\ \cline{1-4}
\multicolumn{4}{|l|}{\textbf{Gender:}}                                                                                                                                                                                                                                                                                                 &  \\ \cline{1-4}
\multicolumn{1}{|l|}{Male}                                     & \multicolumn{1}{l|}{3,077 (56.9\%)}                                                                         & \multicolumn{1}{l|}{3,382 (56.6\%)}                                                                           & \multicolumn{1}{l|}{0.700}              &  \\ \cline{1-4}
\multicolumn{1}{|l|}{Female}                                   & \multicolumn{1}{l|}{2,326 (43.1\%)}                                                                         & \multicolumn{1}{l|}{2,598 (43.4\%)}                                                                           & \multicolumn{1}{l|}{0.685}              &  \\ \cline{1-4}
\multicolumn{4}{l}{}                                                                                                                                                                                                                                                                                                                   &  \\ \cline{1-4}
\multicolumn{1}{|l|}{\textbf{Body Mass Index (BMI)}}           & \multicolumn{1}{l|}{28.7 / 27.3}                                                                            & \multicolumn{1}{l|}{28.9 / 27.8}                                                                              & \multicolumn{1}{l|}{0.010}              &  \\ \cline{1-4}
\multicolumn{1}{|l|}{\textbf{Age}}                             & \multicolumn{1}{l|}{65.5 / 67.0}                                                                            & \multicolumn{1}{l|}{65.7 / 67.0}                                                                              & \multicolumn{1}{l|}{0.620}              &  \\ \cline{1-4}
\multicolumn{4}{l}{}                                                                                                                                                                                                                                                                                                                   &  \\ \cline{1-4}
\multicolumn{4}{|l|}{\textbf{Patient Origin:}}                                                                                                                                                                                                                                                                                         &  \\ \cline{1-4}
\multicolumn{1}{|l|}{Emergency Department (ED)}                & \multicolumn{1}{l|}{1,882 (34.8\%)}                                                                         & \multicolumn{1}{l|}{2,824 (47.2\%)}                                                                           & \multicolumn{1}{l|}{\textless 2.2e-16*} &  \\ \cline{1-4}
\multicolumn{4}{l}{}                                                                                                                                                                                                                                                                                                                   &  \\ \cline{1-4}
\multicolumn{4}{|l|}{\textbf{Therapies in ICU:}}                                                                                                                                                                                                                                                                                       &  \\ \cline{1-4}
\multicolumn{1}{|l|}{Mechanical Ventilation}                   & \multicolumn{1}{l|}{1,146 (21.2\%)}                                                                         & \multicolumn{1}{l|}{1,410 (23.6\%)}                                                                           & \multicolumn{1}{l|}{0.003}              &  \\ \cline{1-4}
\multicolumn{1}{|l|}{Vasopressors}                             & \multicolumn{1}{l|}{703 (13.0\%)}                                                                           & \multicolumn{1}{l|}{1,111 (18.6\%)}                                                                           & \multicolumn{1}{l|}{6.6e-16*}           &  \\ \cline{1-4}
\multicolumn{1}{|l|}{Renal Replacement Therapies (RRT)}        & \multicolumn{1}{l|}{215 (4.0\%)}                                                                            & \multicolumn{1}{l|}{66 (1.1\%)}                                                                               & \multicolumn{1}{l|}{\textless 2.2e-16*} &  \\ \cline{1-4}
                                                               &                                                                                                             &                                                                                                               &                                         &  \\ \cline{1-4}
\multicolumn{4}{|l|}{\textbf{Length of Stay (LOS):}}                                                                                                                                                                                                                                                                                   &  \\ \cline{1-4}
\multicolumn{1}{|l|}{ICU LOS, days}                            & \multicolumn{1}{l|}{3.7 / 2.2}                                                                              & \multicolumn{1}{l|}{3.5 / 2.0}                                                                                & \multicolumn{1}{l|}{6.2e-05*}           &  \\ \cline{1-4}
\multicolumn{1}{|l|}{Hospital LOS, days}                       & \multicolumn{1}{l|}{17.9 / 14.6}                                                                            & \multicolumn{1}{l|}{20.2 / 15.0}                                                                              & \multicolumn{1}{l|}{9.5e-06*}           &  \\ \cline{1-4}
\multicolumn{4}{l}{}                                                                                                                                                                                                                                                                                                                   &  \\ \cline{1-4}
\multicolumn{1}{|l|}{\textbf{Hospital Mortality}}              & \multicolumn{1}{l|}{882 (16.3\%)}                                                                           & \multicolumn{1}{l|}{1,089 (18.2\%)}                                                                           & \multicolumn{1}{l|}{0.009}              &  \\ \cline{1-4}
\multicolumn{4}{l}{}                                                                                                                                                                                                                                                                                                                   &  \\ \cline{1-4}
\multicolumn{4}{|l|}{\textbf{Discharge destination:}}                                                                                                                                                                                                                                                                                  &  \\ \cline{1-4}
\multicolumn{1}{|l|}{Home}                                     & \multicolumn{1}{l|}{2,055 (38.0\%)}                                                                         & \multicolumn{1}{l|}{631 (10.6\%)}                                                                             & \multicolumn{1}{l|}{\textless 2.2e-16*} &  \\ \cline{1-4}
\multicolumn{1}{|l|}{Care Facility (Rehabilitation)}           & \multicolumn{1}{l|}{407 (7.5\%)}                                                                            & \multicolumn{1}{l|}{697 (11.7\%)}                                                                             & \multicolumn{1}{l|}{1.5e-13*}           &  \\ \cline{1-4}
\multicolumn{1}{|l|}{Care Facility (Skilled Nursing Facility)} & \multicolumn{1}{l|}{1,052 (19.5\%)}                                                                         & \multicolumn{1}{l|}{1,132 (18.9\%)}                                                                           & \multicolumn{1}{l|}{0.479}              &  \\ \cline{1-4}
$\star$ Presence of statistical significance.
\end{tabular}
\end{table}

\subsection{Machine Learning Validation}

The statistical analysis provided an insightful characterisation of the readmission group, contrasting it with the non-readmission group and across populations. With this information, we can now move to a machine learning analysis and validation.

Table S5 shows the 47 selected features by the greedy feature selection approach using a Random Forest classifier and considering a balanced readmission dataset. The details of these variables representing machine learning features can be found in Table S2. Although we employed several different undersampling techniques to tackle the high level of readmission imbalance, using a balanced readmission rate presented itself as the most successful technique, which was then applied in both eICU and MIMIC IV cohorts at first. Hyper-parameter optimisation was also applied after feature selection, with the aim of improving the predictive performances of the resultant models (Table S4).

\begin{figure}
    \centering
    \includegraphics[scale=1.0]{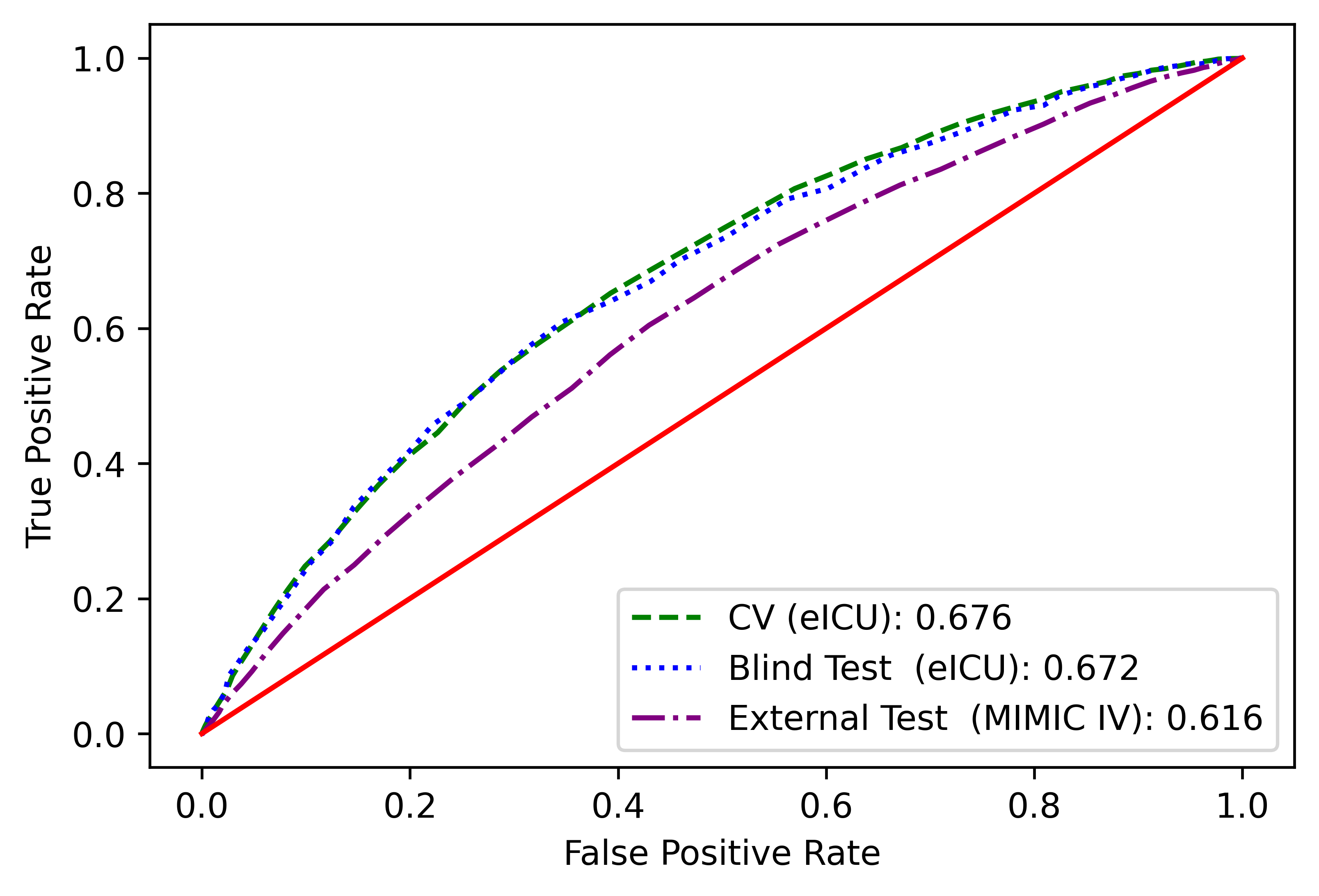}
    \caption{The predictive performance of the proposed readmission model on 10-fold cross-validation and blind test on eICU data. External validation was made utilising MIMIC IV data.
}
    \label{fig_2}
\end{figure}

When bringing all these ML components together, our proposed pipeline achieved an area under the ROC curve (AUC) of 0.68 on multicentric eICU data under a 10-fold cross-validation procedure with a Random Forest classifier with 80 trees in a balanced readmission scenario. It is worth noting that this classification model was considered the best against 9680 other machine learning models in terms of a trade-off among AUC, MCC and recall metrics on 10-fold cross-validation. Explainability was also a key point in selecting the model.  Therefore, the selected variables (features) were taken into consideration while selecting the predictive model itself. On the blind test (stratified 10\% from eICU data), which internally validates the proposed model, consistent results were reached by the predictive model (AUC of 0.672) when compared to 10-fold cross-validation. When externally validated on MIMIC IV's data, our model reached an AUC of 0.616, which demonstrates the overall generalisation capabilities of the proposed model to predict 30-day readmission. Figure \ref{fig_2} shows the AUC plots for these three validation schemes. The results for other performance measures are also summarised in Table S6.

\subsection{Calibration and Likelihood Analysis}

Based on the performed Calibration Analysis (Supplementary Results, Table S7 and Figures S4-S6), the calibration curves, defined from our 30-day ICU readmission model during cross-validation, blind testing and external testing,  indicated slight overestimation/underestimation. Therefore, our main conclusion is that our model is well-calibrated in all sets, demonstrating its clinical usability in an ICU setting. 

\subsection{Likelihood Ratio Analysis}

In addition, our 30-day ICU readmission model faced an analysis based on Likelihood Ratios (see  Supplementary Results and Figure S7), which highlighted the importance of properly setting the classification probability threshold. This analysis step is beneficial for having a better diagnostic readmission profile for ICU patients. In summary, the ICU predictive model's diagnostic power depends on a proper analysis of these thresholds, which was properly defined after the Likelihood Ratio Analysis (Supplementary Results).

\subsection{Explainable Machine Learning for ICU Readmission}

\begin{figure}
    \centering
    \includegraphics[scale=0.65]{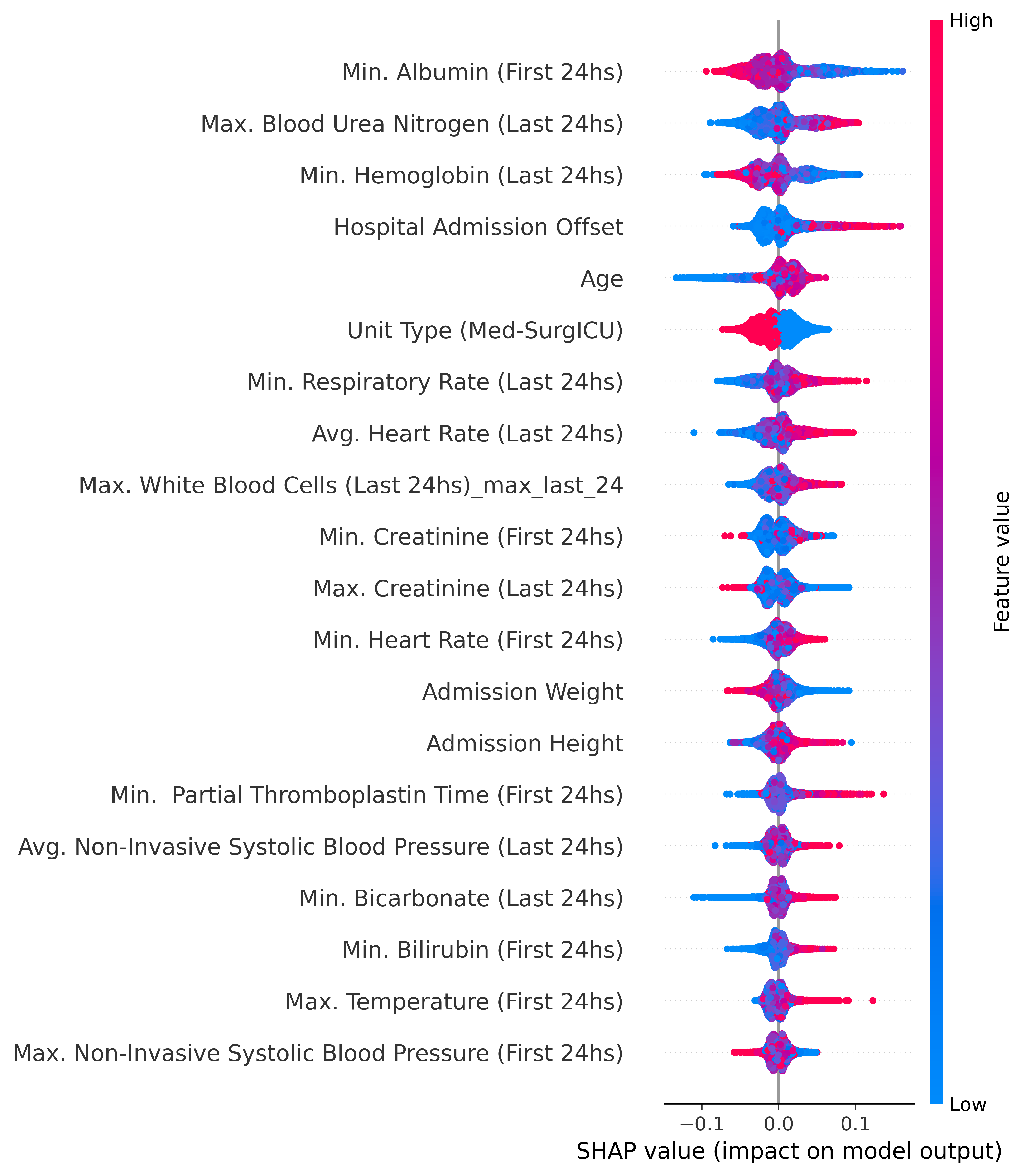}
    \caption{The SHAP summary plot for our proposed readmission model on eICU training data. We show the 20 features with higher SHAP values, i.e., that have a higher impact on the model’s predictive outputs.}
    \label{fig_3}
\end{figure}

We used SHAP to demonstrate how individual features included in the model (see Tables S2 and S5) influence overall readmission predictions. Figure \ref{fig_3} summarises and highlights the 20 most important features, based on SHAP values, we employed to characterise ICU readmission in eICU and MIMIC IV.

Most of the features in the SHAP plot of Figure \ref{fig_3} are vital sign- and blood test-related (see Table S2).  The three most important in this category are the minimum value of albumin in the blood during the first 24 hours, the maximum level of blood urea nitrogen (BUN) during the last 24 hours and the minimum level of hemoglobin during the last 24 hours. 

We identified through the SHAP tree explanation model that high values for the variables Min. Albumin (First 24hs) and Min. Hemoglobin (Last 24hs) are usually more linked to the non-readmission of patients, while low values for them are more present in readmitted patients. Max. Blood Urea Nitrogen (Last 24hs) reveals an opposite trend, where higher values for this variable are fairly more associated with ICU patient readmission, and there is clear evidence that low values for the Max. Blood Urea Nitrogen (Last 24hs) result are attached to non-readmission.

\section{Discussion}

Although it is challenging to map and match common variables or characteristics across different cohorts, our study drove important clinical insights while analysing eICU and comparing it to MIMIC IV (Section 3.1). We discovered how heterogeneous the readmission monocentric and multicentric populations are in general. For example, the readmitted patients in MIMIC IV were treated more frequently with ventilation and vasopressors, although renal replacement therapies were more common in patients readmitted to eICU. Such contrasts in the cohort populations highlight the ability of our standardised pipeline to deal with heterogeneous readmission data.

Our proposed machine learning pipeline learned proper patterns on multicentric ICU data, consequently generalising well on an independent blind test set over the same data (Section~3.2). Our proposed machine learning model captured the essence of readmission, being able to transfer similar predictive performance to external validation on monocentric ICU data (Section~3.2). Overall, this shows that learning from data coming from multiple ICUs has its limitations but may lead to generalisable predictive performance if done adequately. 

A reasonable calibration level is also an essential aspect of the proposed readmission model (Supplementary Results). As a result, our proposed model shares meaningful clinical decision-making due to its ability to calculate individualised probabilistic estimates of readmission \cite{Munn2022}. For example, clinicians may decide that patients with a risk of readmission at least double that of the average risk in their ICU population should be flagged for routine monitoring following discharge from ICU. If the patient is moved to the ward, this may take the form of a brief clinical review by an intensivist each day following transfer to identify early signs of clinical deterioration. This could potentially facilitate the escalation of ward-based measures to optimise the patient’s management and avoid ICU readmission, or it could facilitate better preparation and planning such that unplanned readmissions are minimised.

A good diagnostic level, when we increase the classification probability threshold, complements the model’s predictive performance (Supplementary Results). At the threshold(s) of 0.712 (and 0.813), our model returns the most reliable likelihood ratio among cross-validation, blind testing and external validation, yielding trustworthy readmission predictions. 

Finally, we highlight how the 20 most impactful features are linked with readmission prediction through their explainability (Section 3.3 and Supplementary Results) \cite{Rojas2018,Herrmann1992,Arihan2018,Li2021,Hammer2021,Warner2022,Kerfeld2020,Lee2022,Glans2020,Pedersen2017,Howie2019,You2013,Mishra2021}. Overall, this shows our proposed model was able to learn and capture good insights on eICU data, being compatible with results and discussion in the ICU readmission literature. Supplementary Discussion provides more details on the model's explainability, strength and limitations of this work and its relationships to other studies.

\section*{Authors' contributions}

\textbf{Alex G. C. de Sá:} Conceptualisation; Formal analysis; Investigation; Methodology; Project administration; Resources; Software; Validation; Visualisation; Writing – original draft; Writing – review \& editing.

\textbf{Daniel Gould:}  Conceptualisation; Formal analysis; Investigation; Methodology; Project administration; Validation; Visualisation; Writing – original draft; Writing – review \& editing.

\textbf{Anna Fedyukova:} Data curation; Formal analysis; Methodology; Software; Validation
Visualisation; Writing – original draft.

\textbf{Mitchell Nicholas:} Validation; Writing – review \& editing.

\textbf{Calvin Fletcher:} Validation; Writing – original draft; Writing – review \& editing.

\textbf{Lucy Dockrell:} Validation; Writing – review \& editing.

\textbf{David Pilcher:}   Supervision; Validation; Writing – review \& editing.

\textbf{Daniel Capurro}: Conceptualisation; Validation; Writing – review \& editing.

\textbf{David B. Ascher:}  Supervision; Validation; Writing – review \& editing.

\textbf{Khaled El-Khawas:}  Conceptualisation; Investigation; Methodology; Project administration; Supervision; Validation; Writing – original draft; Writing – review \& editing.

\textbf{Douglas E. V. Pires:}  Conceptualisation; Methodology; Project administration; Supervision
Validation; Writing – original draft; Writing – review \& editing.

\section*{Acknowledgements}
D.B. A. was supported by an Investigator Grant from the National Health and Medical Research Council (NHMRC) of Australia (GNT1174405 to D.A.). Supported in part by the Victorian Government’s Operational Infrastructure Support Program.

\section*{Statement on Conflicts of Interest}
None declared.

\section*{Summary Table}
\begin{itemize}
\item We propose a standardised and explainable machine learning pipeline for 30-day readmission prediction in multicentric eICU and monocentric MIMIC IV cohorts.
\item Our random forest-based readmission model yields overall good calibration and generalisable predictive performance on internal and external validation sets. 
\item Calibration and likelihood ratio analysis also demonstrated the proper clinical (diagnostic) usability of our proposed model, even with the challenging nature of the studied readmission datasets.
\item The derived explanations from the proposed readmission model provide invaluable guidelines, which might be helpful for ICU clinicians’ decision-making while discharging patients.
\end{itemize}

\section*{Supplementary Information}
Supplementary Materials are available at \href{https://tinyurl.com/fne6xv2t}{https://tinyurl.com/fne6xv2t}.

\bibliographystyle{elsarticle-num-names}
\bibliography{references.bib}

\end{document}